\PassOptionsToPackage{unicode}{hyperref}
\PassOptionsToPackage{hyphens}{url}
\PassOptionsToPackage{dvipsnames,svgnames,x11names}{xcolor}
\documentclass[
]{article}
\usepackage{xcolor}
\usepackage[top=10pc,bottom=10pc,left=11pc,right=11pc,heightrounded]{geometry}
\usepackage{amsmath,amssymb}
\usepackage[T1]{fontenc}
\usepackage[utf8]{inputenc}
\usepackage{textcomp}
\usepackage{lmodern}
\IfFileExists{upquote.sty}{\usepackage{upquote}}{}
\IfFileExists{microtype.sty}{
  \usepackage[]{microtype}
  \UseMicrotypeSet[protrusion]{basicmath} 
}{}

\usepackage{longtable,booktabs,array}
\usepackage{calc} 
\usepackage{etoolbox}
\makeatletter
\patchcmd\longtable{\par}{\if@noskipsec\mbox{}\fi\par}{}{}
\makeatother
\IfFileExists{footnotehyper.sty}{\usepackage{footnotehyper}}{\usepackage{footnote}}
\makesavenoteenv{longtable}
\usepackage{graphicx}
\makeatletter
\newsavebox\pandoc@box
\newcommand*\pandocbounded[1]{
  \sbox\pandoc@box{#1}%
  \Gscale@div\@tempa{\textheight}{\dimexpr\ht\pandoc@box+\dp\pandoc@box\relax}%
  \Gscale@div\@tempb{\linewidth}{\wd\pandoc@box}%
  \ifdim\@tempb\p@<\@tempa\p@\let\@tempa\@tempb\fi
  \ifdim\@tempa\p@<\p@\scalebox{\@tempa}{\usebox\pandoc@box}%
  \else\usebox{\pandoc@box}%
  \fi%
}
\def\fps@figure{htbp}
\makeatother

\NewDocumentCommand\citeproctext{}{}
\NewDocumentCommand\citeproc{mm}{%
  \begingroup\def\citeproctext{#2}\cite{#1}\endgroup}
\makeatletter
 \let\@cite@ofmt\@firstofone
 \def\@biblabel#1{}
 \def\@cite#1#2{{#1\if@tempswa , #2\fi}}
\makeatother
\newlength{\cslhangindent}
\newlength{\csllabelwidth}
\newenvironment{CSLReferences}[2] 
 {\begin{list}{}{%
  \setlength{\itemindent}{0pt}
  \setlength{\leftmargin}{0pt}
  \setlength{\parsep}{0pt}
  \ifodd #1
   \setlength{\leftmargin}{\cslhangindent}
   \setlength{\itemindent}{-1\cslhangindent}
  \fi
  \setlength{\itemsep}{#2\baselineskip}}}
 {\end{list}}
\usepackage{calc}

\providecommand{\tightlist}{%
  \setlength{\itemsep}{0pt}\setlength{\parskip}{0pt}}

\usepackage[all,defaultlines=2]{nowidow}

\usepackage{enumitem}
\setlist[1]{labelindent=\parindent}
\setlist[itemize]{leftmargin=*}
\setlist[enumerate]{leftmargin=*}

\setlist[description]{style=unboxed}

\usepackage[titles]{tocloft}

\AtBeginEnvironment{longtable}{\setlist[itemize]{nosep, wide=0pt, leftmargin=*, before=\vspace*{-\baselineskip}, after=\vspace*{-\baselineskip}}}

\usepackage{setspace}

\usepackage[overload]{textcase}
\usepackage[runin]{abstract}

\abslabeldelim{\quad}
\newenvironment{keywords}
{\vskip -3em \hspace{\parindent}\small\sffamily{\sffamily\footnotesize\bfseries\MakeUppercase{Keywords}}\quad}
{\vskip 3em}

\usepackage{titling}
\pretitle{\par\vskip 5em \begin{flushleft}\LARGE\sffamily\bfseries}
\posttitle{\par\end{flushleft}\vskip 0.75em}

\renewcommand{\and}{\end{tabular} \hskip 3em \begin{tabular}[t]{@{\hspace{0em}}l@{}}}
\preauthor{\begin{flushleft}
           \lineskip 1.5em 
           \begin{tabular}[t]{@{\hspace{0em}}l@{}}}
\postauthor{\end{tabular}\par\end{flushleft}}

\predate{}
\postdate{}

\newcommand{\published}[1]{%
   \gdef\puB{#1}}
   \newcommand{\puB}{}

\usepackage{titlesec}
\titleformat*{\section}{\Large\sffamily\bfseries\raggedright}
\titleformat*{\subsection}{\large\sffamily\bfseries\raggedright}
\titleformat*{\subsubsection}{\normalsize\sffamily\bfseries\raggedright}
\titleformat*{\paragraph}{\small\sffamily\bfseries\raggedright}

\titlespacing*{\section}{0em}{2em}{0.1em}
\titlespacing*{\subsection}{0em}{1.25em}{0.1em}
\titlespacing*{\subsubsection}{0em}{0.75em}{0em}

\usepackage[bottom, flushmargin]{footmisc}

\usepackage[font={small,sf}, labelfont={small,sf,bf}]{caption}

\usepackage{pdflscape}
\newcommand{\blandscape}{\begin{landscape}}
\newcommand{\elandscape}{\end{landscape}}

\makeatletter
\@ifpackageloaded{caption}{}{\usepackage{caption}}
\AtBeginDocument{%
\ifdefined\contentsname
  \renewcommand*\contentsname{Table of contents}
\else
  \newcommand\contentsname{Table of contents}
\fi
\ifdefined\listfigurename
  \renewcommand*\listfigurename{List of Figures}
\else
  \newcommand\listfigurename{List of Figures}
\fi
\ifdefined\listtablename
  \renewcommand*\listtablename{List of Tables}
\else
  \newcommand\listtablename{List of Tables}
\fi
\ifdefined\figurename
  \renewcommand*\figurename{Figure}
\else
  \newcommand\figurename{Figure}
\fi
\ifdefined\tablename
  \renewcommand*\tablename{Table}
\else
  \newcommand\tablename{Table}
\fi
}
\@ifpackageloaded{float}{}{\usepackage{float}}
\floatstyle{ruled}
\@ifundefined{c@chapter}{\newfloat{codelisting}{h}{lop}}{\newfloat{codelisting}{h}{lop}[chapter]}
\floatname{codelisting}{Listing}

\makeatother
\makeatletter
\@ifpackageloaded{caption}{}{\usepackage{caption}}
\@ifpackageloaded{subcaption}{}{\usepackage{subcaption}}
\makeatother
\usepackage{bookmark}
\IfFileExists{xurl.sty}{\usepackage{xurl}}{} 
\hypersetup{
  pdftitle={Meet Your New Client: Writing Reports for AI -- Benchmarking Information Loss in Market Research Deliverables},
  pdfauthor={Paul F. Simmering; Benedikt Schulz; Oliver Tabino; Dr.~Georg Wittenburg},
  pdfkeywords={retrieval-augmented generation, knowledge
management, market research, large language models},
  colorlinks=true,
  linkcolor={DarkSlateBlue},
  filecolor={Maroon},
  citecolor={DarkSlateBlue},
  urlcolor={DarkSlateBlue},
  pdfcreator={LaTeX via pandoc}}

\usepackage{orcidlink}  

\title{Meet Your New Client: Writing Reports for AI -- Benchmarking
Information Loss in Market Research Deliverables\thanks{We would like to
thank the participants of the AI Forum of the German Society for Online
Research (DGOF) for the discussion about knowledge management that
inspired this study. We are also grateful for helpful comments by
participants at the General Online Research 2025 conference. Paul
Simmering acknowledges inference credits granted by Fireworks AI and
compute credits granted by Modal.}}

\author{
{\large Paul F. Simmering}%
\thanks{Corresponding author.} \\%
Q Agentur für Forschung GmbH \\%
{\footnotesize \url{paul@simmering.dev}} \and
{\large Benedikt Schulz}%
 \\%
Q Agentur für Forschung GmbH \\%
{\footnotesize \url{benedikt.ps@gmail.com}} \and
{\large Oliver Tabino}%
 \\%
Q Agentur für Forschung GmbH \\%
{\footnotesize \url{oliver.tabino@teamq.de}} \and
{\large Dr.~Georg Wittenburg}%
 \\%
Inspirient GmbH \\%
{\footnotesize \url{georg.wittenburg@inspirient.com}} \and
}

\date{}

\usepackage{url}
\begin{document}
\published{\textbf{2025-08-13} \\ {\scriptsize Access the code and data
at \url{https://github.com/psimm/meet-your-new-client}}}

\maketitle

\begin{abstract}
As organizations adopt retrieval-augmented generation (RAG) for their
knowledge management systems (KMS), traditional market research
deliverables face new functional demands. While PDF reports and slides
have long served human readers, they are now also ``read'' by AI systems
to answer user questions. To future-proof reports being delivered today,
this study evaluates information loss during their ingestion into RAG
systems. It compares how well PDF and PowerPoint (PPTX) documents
converted to Markdown can be used by an LLM to answer factual questions
in an end-to-end benchmark. Findings show that while text is reliably
extracted, significant information is lost from complex objects like
charts and diagrams. This suggests a need for specialized, AI-native
deliverables to ensure research insights are not lost in translation.
\end{abstract}
\vskip 3em

\begin{keywords}
\def\sep{;\ }
retrieval-augmented generation\sep knowledge management\sep market
research\sep 
large language models
\end{keywords}


\section{Introduction}\label{introduction}

As organizations implement retrieval-augmented generation (RAG) in
knowledge management systems (KMS), the audience for market research
deliverables is shifting. Reports in PDF and PowerPoint (PPTX) formats,
once designed for human readers, are now also consumed by AI systems.
For consultancies and market research agencies, this means that
effective packaging of results for AI readability is becoming a core
requirement. Information loss during document ingestion can undermine
downstream retrieval and generation, making the choice of format and
structure critical for future-proofing deliverables.

Recent surveys and case studies (e.g., M. Cheng et al.
(\citeproc{ref-cheng2025surveyknowledgeorientedretrievalaugmentedgeneration}{2025}),
Gao et al.
(\citeproc{ref-gao2024retrievalaugmentedgenerationlargelanguage}{2024}),
PricewaterhouseCoopers (\citeproc{ref-PwC2024CaseStudy}{2024}))
highlight the growing adoption of RAG and the importance of document
structure for effective AI use. However, standard delivery formats may
not be optimal for machine processing, especially when information is
embedded in and/or implied via complex layouts or images.

This study quantifies the information loss that occurs when documents
are converted to Markdown for RAG. We systematically compare how well
documents in PDF and PPTX formats can be used by large language models
to answer factual questions. Our end-to-end evaluation uses four
open-source Markdown conversion libraries (Docling, Marker, Markitdown,
Zerox OCR) and three vision and question answering models. We introduce
a new QA benchmark and corpus of 41 PPTX documents, expanding on prior
work focused on PDFs. Our findings inform both practitioners and
researchers on the limitations of current formats and tools and provide
concrete guidance for agencies seeking to deliver AI-ready reports.

\section{Related work}\label{related-work}

Research on the topic comes from two directions: (1) document layout
analysis benchmarks for the accuracy of document segmentation and
conversion, and (2) question-answering benchmarks that evaluate the
accuracy of an answer generation model. As an end-to-end benchmark, this
study bridges the gap between the two.

\subsection{Document layout analysis (DLA)
benchmarks}\label{document-layout-analysis-dla-benchmarks}

DLA benchmarks focus on the accuracy of document segmentation using
image models. DocLayNet by Pfitzmann et al.
(\citeproc{ref-pfitzmann2022doclaynet}{2022}) is the largest benchmark
for PDF layout segmentation, featuring 11 layout elements, including
text, tables, and pictures, across 80863 manually annotated pages.
Later, H. Cheng et al.
(\citeproc{ref-chengM6DocLargeScaleMultiFormat2023}{2023}) added
M\textsuperscript{6}Doc, which includes scanned documents, 74
fine-grained layout elements, and a large Chinese corpus.

OmniDocBench by Ouyang et al.
(\citeproc{ref-ouyangOmniDocBenchBenchmarkingDiverse2025}{2025})
introduced PDF slides as a document type, relevant to our study. Going
beyond DLA, it evaluates tools like MinerU, Marker, Mathpix, and VLMs by
comparing their Markdown output to ground truth. Images are removed in
pre-processing. Pipeline tools and VLMs performed comparably, with
MinerU excelling on English and Qwen2-VL-72B by Alibaba on Chinese.
Qwen2-VL-72B also performed best on the JSON extraction OCR benchmark by
OmniAI (\citeproc{ref-omni-benchmarking-ocr-2025}{2025}).

Specialized benchmarks for specific layout elements include PlotQA by
Methani et al. (\citeproc{ref-methani2020plotqa}{2020}) and ChartQA by
Masry et al.
(\citeproc{ref-masry2022chartqabenchmarkquestionanswering}{2022}) for
chart understanding, HybridQA by W. Chen et al.
(\citeproc{ref-chen-etal-2020-hybridqa}{2020}) for table-based QA, and
PubTables-1M by Smock, Pesala, and Abraham
(\citeproc{ref-smock2022pubtables}{2022}) for table extraction from
scientific papers.

\subsection{Question-answering (QA)
benchmarks}\label{question-answering-qa-benchmarks}

Question-answering benchmarks evaluate how reliably models can produce
correct answers from document-derived evidence. Comprehensive surveys
map this space and its task formulations
(\citeproc{ref-cambazoglu2021review}{Cambazoglu et al. 2021};
\citeproc{ref-wang2022modernquestionansweringdatasets}{Wang 2022};
\citeproc{ref-Rogers_2023}{Rogers, Gardner, and Augenstein 2023}). In
the taxonomy of Rogers, Gardner, and Augenstein
(\citeproc{ref-Rogers_2023}{2023}), the present study falls under
probing questions with free-form answers over expert materials in a
monolingual (English) setting. It also works directly with realistic
unstructured documents, rather than starting from cleaned text-only
representations in classic benchmarks such as HotpotQA
(\citeproc{ref-hotpotqa2018}{Z. Yang et al. 2018}).

FinanceBench
(\citeproc{ref-islam2023financebenchnewbenchmarkfinancial}{Islam et al.
2023}) is closest in spirit among end-to-end evaluations over real
documents. It targets financial filings and reports. The benchmark is
limited to text and tabular layout elements, starting from PDFs that are
converted to text using PyMuPDF and Langchain, without image extraction.
A key finding is that a long-context approach, where the full document
text is provided to the model, outperforms vector-store variants. Their
best configuration (long context with GPT‑4 Turbo) reaches 79\%
accuracy. FinanceBench further reports that placing relevant context
before the question improves performance. These observations motivate
our own use of a long-context setting and prompt.

The UDA benchmark (\citeproc{ref-hui2024udabenchmarksuiteretrieval}{Hui,
Lu, and Zhang 2024}) advances realism by operating on raw PDFs and HTML
rather than pre-cleaned text, aggregating diverse datasets and question
types. It concentrates on text and tables and compares several
extraction and parsing strategies. Notably, a VLM--based table parsing
approach outperforms classic extraction, and a sequential ``parse first,
then answer'' pipeline can surpass direct image-questioning. UDA also
contrasts long-context with retrieval pipelines, reporting broadly
similar performance on knowledge tasks but and advantage for RAG on
numerical reasoning. While UDA omits data charts, diagrams, and images
as layout elements, its approach and focus on real-world usefulness
motivate our study.

Building on these lines of work, we extend end-to-end QA to PPTX and to
diagrams, data charts, and images. We foreground the conversion step,
comparing four Markdown conversion libraries, and quantify how format
(PDF vs PPTX) and layout affect QA accuracy. The emphasis is practical:
guidance for market research teams on format and layout choices that
preserve answerable content in KMS.

\section{Benchmark}\label{benchmark}

\subsection{Overview}\label{overview}

\begin{figure}

\centering{

\pandocbounded{\includegraphics[keepaspectratio]{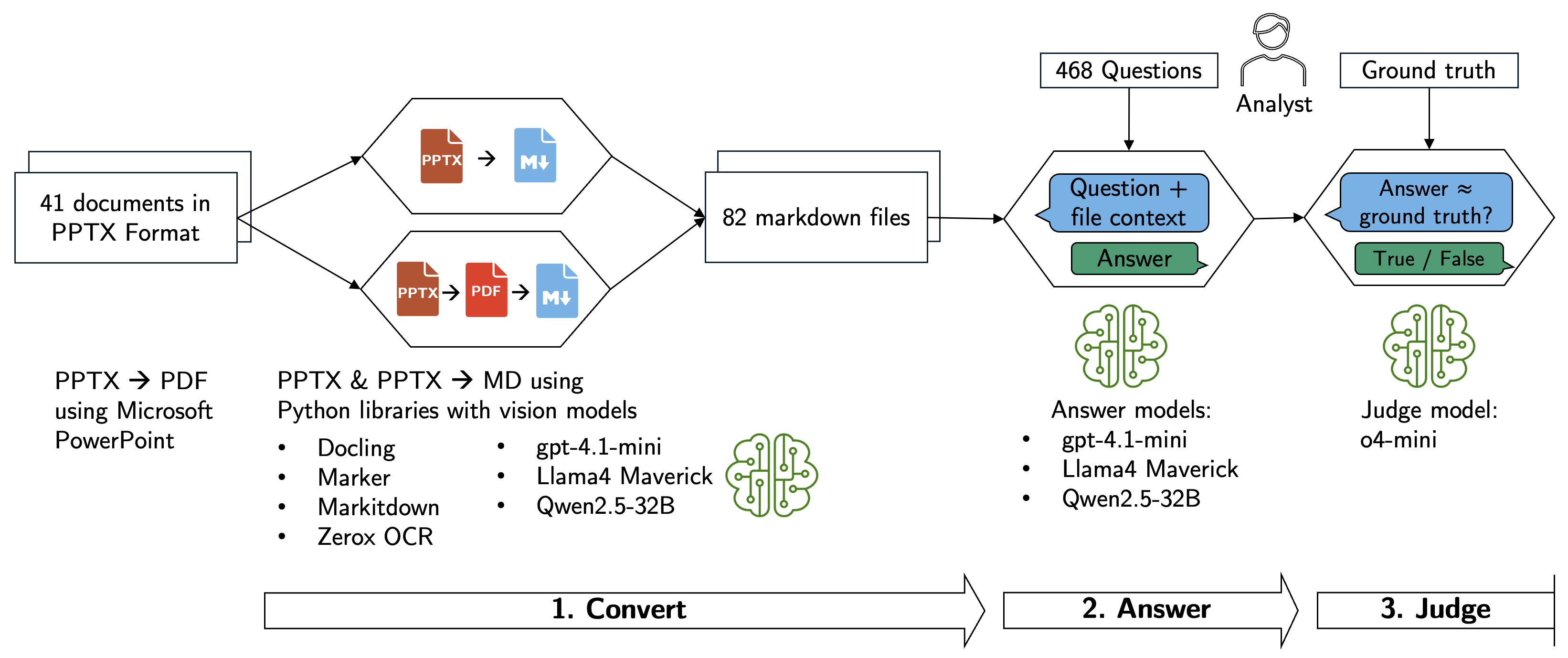}}

}

\caption{\label{fig-overview}Experiment setup diagram.}

\end{figure}

Figure~\ref{fig-overview} illustrates the experiment setup. It consists
of 468 questions relating to 41 PPTX documents. Each document is
manually converted into PDF format using PowerPoint. Then, the PPTX and
PDF versions of the documents are each converted to Markdown via
conversion libraries (see Section~\ref{sec-conversion-libraries}). This
results in two Markdown files per original document. These form the
knowledge base for the question answering task. A language model is
prompted with each question and given the corresponding document's full
Markdown text as context. Finally, the answers are compared to the
human-annotated ground truth by an LLM as a judge.

\subsection{File formats}\label{file-formats}

A key difference between the file formats lies in their structure.
Livathinos et al.
(\citeproc{ref-livathinos2025doclingefficientopensourcetoolkit}{2025})
categorize the formats into low-level formats (PDF and PNG) and
markup-based formats (PPTX).

\begin{itemize}
\tightlist
\item
  \textbf{PPTX} is a semantic, XML-based format that stores objects like
  text and tables in a structured hierarchy. In theory, this structure
  is highly amenable to machine processing. Images are an exception and
  require a VLM to be captioned.
\item
  \textbf{PDF} is a visual format describing page layout, not semantic
  content. Reconstructing the document's structure from low-level
  drawing instructions is complex and prone to information loss.
\end{itemize}

Each step involving a VLM or other stochastic model carries a risk of
information loss or hallucination. Based on the differences in
structure, a starting hypothesis is that PPTX files will be converted to
Markdown with less information loss than PDF files.

\subsection{Collection of documents}\label{collection-of-documents}

The PPTX documents were sourced from publicly accessible web sources and
additional documents were supplied by Q Agentur für Forschung, a market
research agency. The majority of the files were found in GitHub
repositories. These were found via the GitHub API and then manually
inspected. Only English documents that contain information across at
least three different layout elements were included. Document length
ranges from 8 to 106 slides, with an average of 38.3 slides per
document.

In regard to content, the documents are market research reports,
conference presentations and lecture slides. They cover a range of
topics, from cosmetics trends to political science. Due to the nature of
GitHub as a source, computer science is the most common topic. A
complete list of the documents, including their sources and licenses, is
available in the GitHub repository.

\subsection{Question formulation}\label{question-formulation}

Each item in the benchmark is a triplet (\(D\), \(q\), \(a\)), where
\(D\) is a PPTX document in its original format, \(q\) is the question
answerable using information in \(D\), and \(a\) is the ground truth
answer to the question. This mirrors the format of UDA benchmarks
(\citeproc{ref-hui2024udabenchmarksuiteretrieval}{Hui, Lu, and Zhang
2024}). A total of 468 such questions were formulated across the 41
documents, amounting to an average of 11.4 questions per document.

The content that questions relate to is distributed across the slides.
Some relate to the first slide, others up to slide 99. The average
content position is at 14.9 slides.

The questions were manually formulated by two of the authors. When
formulating the questions, care was taken to ensure that they could be
answered unambiguously. To isolate the impact of different formats, each
question was designed to be answerable using information from a single
slide and a single layout element. However, information can be present
in multiple forms in the same document. For example, a text box can
repeat information that is also present in a data chart or a table. When
asked a question about the report, the model could use either layout
element or a combination to find the answer. It may also infer the
correct answer from the surrounding text. As Rogers, Gardner, and
Augenstein (\citeproc{ref-Rogers_2023}{2023}) note ``we need to
reconsider the idea that whether or not a given reasoning skill is
`required' is a characteristic of a given question. It is rather a
characteristic of the combination of that question and the entire
dataset''. In some instances, text boxes serve as axis labels for a data
chart, requiring the model to understand the relationship between the
chart and the text.

\subsection{Layout elements}\label{layout-elements}

\begin{figure}

\centering{

\includegraphics[width=1\linewidth,height=\textheight,keepaspectratio]{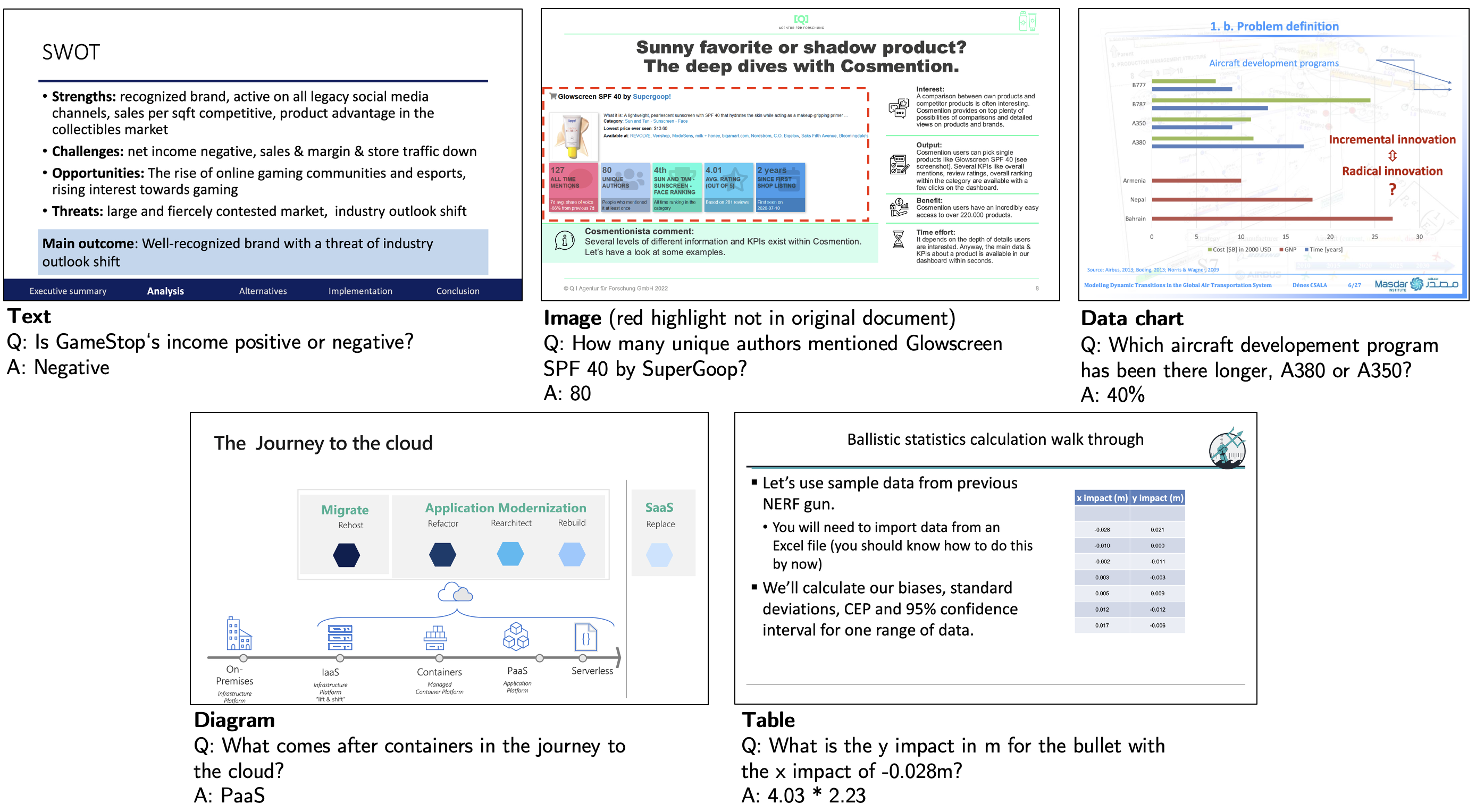}

}

\caption{\label{fig-data-types}Example slides, questions and answers for
the five layout elements.}

\end{figure}

Questions pertain to information presented in different layout elements.
We distinguish between five layout elements found in PowerPoint files,
shown in Figure~\ref{fig-data-types}.

\begin{itemize}
\tightlist
\item
  \textbf{Text}: Information within a single, contiguous text field.
\item
  \textbf{Image}: Visual elements, including photographs and screenshots
  of other layout elements such as a table saved as an image.
\item
  \textbf{Table}: Structured data arranged in rows and columns.
\item
  \textbf{Diagram}: Combinations of text and graphical elements, such as
  flow charts.
\item
  \textbf{Data Chart}: Graphical representations of numeric data, such
  as bar or line charts. Only native PowerPoint charts are included;
  charts inserted as images are classified as \emph{Image}.
\end{itemize}

Other data in PowerPoint, such as videos and speaker notes, are outside
of the scope of this study.

To avoid an outsized impact from a single unusually easy or difficult
object on the result, a maximum of three questions were asked regarding
one layout element, for example a diagram on a particular slide.

\subsection{Document conversion
libraries}\label{sec-conversion-libraries}

A common entry point into a text-based RAG system is to convert
documents to Markdown. The plain text format is token-efficient,
human-readable, and supports many formatting options. Images are
supported as external references with optional captions. LLMs can also
generate Markdown themself
(\citeproc{ref-chenMDEvalEvaluatingEnhancing2025}{Z. Chen et al. 2025}),
making it useful for summaries and intermediate outputs. The two main
conversion paradigms are (1) pipeline-based decomposition using
specialized object detection models and (2) end-to-end processing using
a vision language model (VLM).

Table~\ref{tbl-conversion-libraries} gives an overview of open-source
libraries to convert PDF and PPTX to Markdown. In addition, there are
proprietary solutions like LlamaParse and Mathpix. The pipeline library
\href{https://github.com/Unstructured-IO/unstructured}{unstructured} is
omitted as it does not support export to Markdown at the time of
writing. PyMuPDF and pypdf are included because they are used in related
benchmarks, however, they do not support PPTX conversion. MinerU does
not support PPTX files, but the wrapper library magic-doc does. However,
the version of MinerU supported by magic-doc lags behind the current
version of MinerU.

\begin{longtable}[]{@{}
  >{\raggedright\arraybackslash}p{(\linewidth - 10\tabcolsep) * \real{0.2174}}
  >{\raggedright\arraybackslash}p{(\linewidth - 10\tabcolsep) * \real{0.2609}}
  >{\raggedright\arraybackslash}p{(\linewidth - 10\tabcolsep) * \real{0.1304}}
  >{\raggedright\arraybackslash}p{(\linewidth - 10\tabcolsep) * \real{0.0870}}
  >{\raggedright\arraybackslash}p{(\linewidth - 10\tabcolsep) * \real{0.0870}}
  >{\raggedright\arraybackslash}p{(\linewidth - 10\tabcolsep) * \real{0.2174}}@{}}
\toprule\noalign{}
\begin{minipage}[b]{\linewidth}\raggedright
Library
\end{minipage} & \begin{minipage}[b]{\linewidth}\raggedright
Developer
\end{minipage} & \begin{minipage}[b]{\linewidth}\raggedright
Paradigm
\end{minipage} & \begin{minipage}[b]{\linewidth}\raggedright
PDF
\end{minipage} & \begin{minipage}[b]{\linewidth}\raggedright
PPTX
\end{minipage} & \begin{minipage}[b]{\linewidth}\raggedright
License
\end{minipage} \\
\midrule\noalign{}
\endfirsthead
\toprule\noalign{}
\begin{minipage}[b]{\linewidth}\raggedright
Library
\end{minipage} & \begin{minipage}[b]{\linewidth}\raggedright
Developer
\end{minipage} & \begin{minipage}[b]{\linewidth}\raggedright
Paradigm
\end{minipage} & \begin{minipage}[b]{\linewidth}\raggedright
PDF
\end{minipage} & \begin{minipage}[b]{\linewidth}\raggedright
PPTX
\end{minipage} & \begin{minipage}[b]{\linewidth}\raggedright
License
\end{minipage} \\
\midrule\noalign{}
\endhead
\bottomrule\noalign{}
\tabularnewline
\caption{Comparison of document conversion libraries. *Marker and Zerox
OCR convert PPTX to PDF at the start of the conversion. **model weights
for Marker are licensed
cc-by-nc-sa-4.0.}\label{tbl-conversion-libraries}\tabularnewline
\endlastfoot
\href{https://github.com/docling-project/docling}{Docling} & IBM &
Pipeline & Yes & Yes & MIT \\
\href{https://github.com/CosmosShadow/gptpdf}{GPTPDF} & Chen Li & VLM &
Yes & No & MIT \\
\href{https://github.com/pymupdf/PyMuPDF}{PyMuPDF} & Artifex Software &
Pipeline & Yes & No & AGPL-3 \\
\href{https://github.com/py-pdf/pypdf}{pypdf} & Martin Thoma & Pipeline
& Yes & No & BSD-3 \\
\href{https://github.com/VikParuchuri/marker}{Marker} & Datalab &
Pipeline & Yes & Yes* & GPL-3** \\
\href{https://github.com/microsoft/markitdown}{Markitdown} & Microsoft &
Pipeline & Yes & Yes & MIT \\
\href{https://github.com/opendatalab/MinerU}{MinerU} & OpenDataLab &
Pipeline & Yes & No & AGPL-3 \\
\href{https://github.com/getomni-ai/zerox}{Zerox OCR} & OmniAI & VLM &
Yes & Yes* & MIT \\
\end{longtable}

This study compares the following four libraries that support both PDF
and PPTX in their current version:

\begin{itemize}
\tightlist
\item
  \emph{Docling} by Livathinos et al.
  (\citeproc{ref-livathinos2025doclingefficientopensourcetoolkit}{2025})
  at IBM. Version 2.38.0 was used (released 2025-06-23).
\item
  \emph{Markitdown} by Microsoft (\citeproc{ref-markitdown}{2024}).
  Version 0.1.2 was used (released 2025-05-28).
\item
  \emph{Marker} by Datalab (\citeproc{ref-marker_github}{2024}). Version
  1.7.5 was used (released 2025-06-11).
\item
  \emph{Zerox OCR} by OmniAI (\citeproc{ref-zerox_github}{2024}).
  Version 0.1.06 was used (released 2024-12-18).
\end{itemize}

All libraries were run with support from a VLM to caption images. The
captions were placed into the Markdown documents at the location of the
images.

\subsection{Conversion library
limitations}\label{sec-conversion-library-limitations}

\textbf{Missing image captions}: Docling does not support image captions
for images in PPTX files. Some images in PDF files are not captioned.
PDF files converted to Markdown by Markitdown exhibit duplicate text and
missing image captions. Markitdown does not support image captions for
images that are in PDF format when they are embedded in PPTX files.

\textbf{Conversion errors}: Marker and the underlying library pillow
were unable to handle Windows Metafile (WMF) files, a legacy format
encountered in 15 benchmarked documents, on the Debian Linux system used
for this study. Markitdown had an issue with 17 PPTX files that contain
shapes with \texttt{None} as top, left, height or width attributes.
These combinations of libraries and documents were excluded from further
analysis.

\subsection{Models}\label{models}

The following LLMs are compared as vision models and as question
answering models:

\begin{itemize}
\tightlist
\item
  \emph{Llama4 Maverick} by Meta (\citeproc{ref-llama4_meta_blog}{Meta
  AI 2025}), hosted on the Fireworks AI API
\item
  \emph{Qwen2.5-32B Instruct} by Alibaba
  (\citeproc{ref-qwenQwen25TechnicalReport2025}{A. Yang et al. 2025}),
  hosted on the Fireworks AI API
\item
  \emph{gpt-4.1-mini-2025-04-14} by OpenAI
  (\citeproc{ref-openai2024gpt4ocard}{OpenAI 2024}), hosted on the
  OpenAI API
\end{itemize}

All of these models have a sufficiently long context length to fit the
longest document in Markdown form. Prompts, temperature, and other token
generation parameters are kept constant. They are documented in the
appendix.

The LLM-as-judge method
(\citeproc{ref-zheng2023judgingllmasajudgemtbenchchatbot}{Zheng et al.
2023}) is employed to evaluate the correctness of the answers. In
contrast to word-based evaluations like BLEU, ROUGE, and METEOR, the
LLM-as-judge method is able to semantically evaluate the answers. It is
not limited to lexical matches. Further, it can be instructed on
specific judging criteria. A reasoning model, \emph{o4-mini-2025-04-16},
carries out the judging in a true-false manner. We provide the prompt in
the appendix. The nature of the questions leaves little room for
interpretation. The sum of correct answers divided by the total number
of questions is the accuracy, the primary metric of this study.

\subsection{Configurations}\label{configurations}

The experiment has a factorial design, combining the four libraries,
each of the three models as vision model for image captioning and as the
question answering model. At the time of writing the Zerox OCR library
was not compatible with Llama4 or Qwen2.5 on hosted inference APIs,
therefore only gpt-4.1-mini was used with it. When Zerox OCR converts
PPTX to Markdown, it starts by converting the PPTX to PDF. This internal
conversion is not the subject of this study, so Zerox is only used with
PDF files. In total, 30 configurations were tested.

\section{Results}\label{results}

\subsection{Layout elements and file
formats}\label{layout-elements-and-file-formats}

\begin{figure}

\centering{

\pandocbounded{\includegraphics[keepaspectratio]{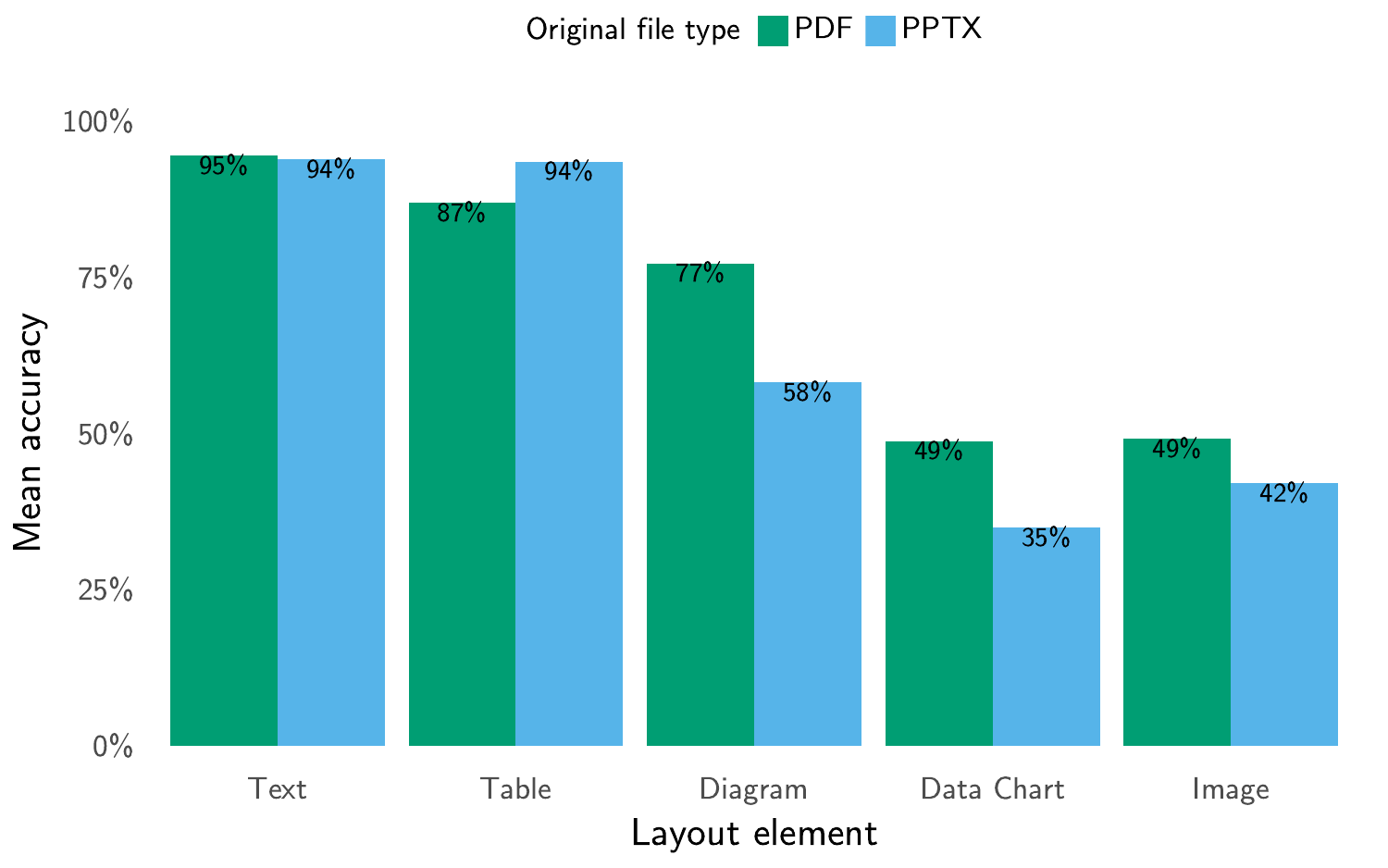}}

}

\caption{\label{fig-acc-by-file-type}Answer accuracy in percent by file
type and layout element. Results are averaged across conversion
libraries, question answering model, and vision model. Conversions that
resulted in errors are excluded.}

\end{figure}%

Figure~\ref{fig-acc-by-file-type} shows the accuracy by file type and
layout element. Table~\ref{tbl-conversion-libraries-acc} shows the same
results split by conversion library. In order of overall accuracy, the
layout elements are:

\begin{enumerate}
\def\labelenumi{\arabic{enumi}.}
\tightlist
\item
  \textbf{Text} is accurately converted by all libraries and with both
  file types, scoring above 90\% accuracy. That makes text the most
  reliable layout element for question answering.
\item
  \textbf{Tables} are accurately converted in all but one configuration,
  with most showing accuracy above 90\%. The exception is Markitdown on
  PDF files, which does not convert tables to Markdown tables. Instead,
  table cells are extracted as text snippets, losing their structural
  relationships.
\item
  \textbf{Diagrams} have lower accuracy than text and tables, with all
  scores below 90\%. Zerox OCR achieves 88.5\% on PDF files, likely due
  to its use of box-drawing characters. Markitdown reaches 80.2\% on
  PPTX files. Other libraries and file type combinations score below
  80\%.
\item
  \textbf{Data charts} display high variance of accuracy across
  libraries and file types. Markitdown excels at this task in PPTX
  files, scoring the highest accuracy of 70.1\%. Accuracy in other
  configurations is low.
\item
  \textbf{Images} show high accuracy variance. Docling cannot caption
  images in PPTX files, and Markitdown cannot in PDF files. Docling
  achieves a comparatively high accuracy of 72.5\% on images in PDF
  files, more than 10 percentage points higher than others using the
  same vision models. This may be due to its option for a custom image
  prompt, which is not available in the other libraries. The image
  prompt is in the appendix.
\end{enumerate}

We do not find support for our hypothesis that PPTX files with their
structured XML format are generally better suited for document
understanding than PDF files. As an external reference to the statistics
reported here, FinanceBench's
(\citeproc{ref-islam2023financebenchnewbenchmarkfinancial}{Islam et al.
2023}) best long-context configuration with GPT4-Turbo reported 79\%
accuracy on questions pertaining to financial PDF documents.

\begin{table}

\centering{

\begin{tabular*}{\linewidth}{@{\extracolsep{\fill}}llrrrrr}
\toprule
 & \multicolumn{5}{c}{Layout element} \\ 
\cmidrule(lr){3-7}
Library & File type & Text & Table & Diagram & Data Chart & Image \\ 
\midrule\addlinespace[2.5pt]
Docling & PDF & 92.9 & 90.4 & 74.3 & 57.1 & 72.5 \\
Docling & PPTX & 94.4 & 94.5 & 60.6 & 16.7 & 9.7 \\
Marker & PDF & 95.8 & 91.2 & 73.5 & 52.9 & 63.8 \\
Marker & PPTX & 93.4 & 94.2 & 54.0 & 18.2 & 57.7 \\
Markitdown & PDF & 94.4 & 77.0 & 80.2 & 32.7 & 8.4 \\
Markitdown & PPTX & 94.0 & 91.9 & 59.9 & 70.1 & 59.0 \\
Zerox OCR & PDF & 96.1 & 94.2 & 88.5 & 60.0 & 58.1 \\
\bottomrule
\end{tabular*}

}

\caption{\label{tbl-conversion-libraries-acc}Answer accuracy in percent
by conversion library, file type and layout element. Results are
averaged across question answering model and vision model.}

\end{table}%

\subsection{Image captioning and question answering
models}\label{image-captioning-and-question-answering-models}

The experiment also involved variations in the used for image captioning
(or alternatively the whole conversion in the case of Zerox OCR) and for
answering questions based on the converted document.
Table~\ref{tbl-model-accuracy} displays accuracy by combination of
models. Out of the three models, GPT 4.1 Mini posts the highest accuracy
scores as a vision model. As question answering models, Llama4 Maverick
and GPT 4.1 Mini reach similar accuracy. The differences illustrate that
every single piece of the pipeline (file type, layout element, vision
model, question answering model, conversion library) influences the
accuracy.

\begin{table}

\centering{

\begin{tabular*}{\linewidth}{@{\extracolsep{\fill}}lrrr}
\toprule
 & \multicolumn{3}{c}{Vision language model} \\ 
\cmidrule(lr){2-4}
Question answering model & Llama4 Maverick & Qwen2.5 32B & GPT 4.1 Mini \\ 
\midrule\addlinespace[2.5pt]
Llama4 Maverick & 67.7 & 67.0 & 72.3 \\
Qwen2.5 32B & 63.9 & 61.7 & 69.6 \\
GPT 4.1 Mini & 67.0 & 65.9 & 72.4 \\
\bottomrule
\end{tabular*}

}

\caption{\label{tbl-model-accuracy}Answer accuracy in percent by QA
model and VLM. Results are averaged across layout elements, file types,
and conversion libraries. Runs with Zerox OCR were excluded, as it is
incompatible with Llama4 and Qwen2.5 on hosted inference APIs.}

\end{table}%

\section{Discussion}\label{discussion}

Findings indicate that the classic formats PPTX and PDF, while valuable
for human interpretation, are not optimal for RAG systems. In some
configurations of libraries, files types and layout elements, answer
accuracy was below 30\%. This is a warning sign for buyers and sellers
of reports about market research and other topics. It is in the interest
of both parties to ensure that information is passed correctly into a
KMS. There are two main paths for this:

\begin{enumerate}
\def\labelenumi{\arabic{enumi}.}
\tightlist
\item
  Coordinate on a file format and layout elements. Ensure that the buyer
  is using a conversion library that is well equipped for the file
  format and the layout elements used in the seller's reports. Adjust
  the reports as necessary to ensure proper comprehension and work
  around edge cases.
\item
  Negotiate a special-purpose deliverable to be sent alongside the main
  report. More on this in
  Section~\ref{sec-special-purpose-deliverables}.
\end{enumerate}

A majority of the tested libraries (Docling, Marker, and Zerox OCR)
prioritize PDF conversion; only Markitdown emphasizes PPTX and other
Microsoft Office formats. This mirrors the relative abundance of PDF
files available on the web, relative to PPTX files. With current tools,
PPTX is less practical than PDF for conversion, despite its foundation
in structured XML.

The tested libraries were not all able to convert each document and
layout element therein. We encountered errors as well as silent
failures. This reflects the libraries' recency and rapid development.
Zerox OCR with its pure vision approach proved to be the most robust.
This success mirrors the results of the table extraction experiment by
Hui, Lu, and Zhang
(\citeproc{ref-hui2024udabenchmarksuiteretrieval}{2024}).

\subsection{Special-purpose
deliverables}\label{sec-special-purpose-deliverables}

Instead of working around the limitations of formats made for human
consumption, a secondary deliverable optimized for AI could be the
solution. Text-heavy reports offer greater compatibility than styled
PPTX and PDF files, yet are not optimal either, for example when
methodology is presented separately from results. Our study suggests
that transitioning to complementary special-purpose deliverables,
designed explicitly for AI, enhances the retrieval accuracy of research
insights within KMS, and thus for the client. Detailed specification of
these special-purpose deliverables is left to future work, but we can
already outline relevant design principles:

\begin{enumerate}
\def\labelenumi{\arabic{enumi}.}
\tightlist
\item
  \textbf{Text-first representation}: Graphical elements should include
  comprehensive textual descriptions capturing both explicit content and
  implicit relationships, including key numbers and trend
  interpretations.
\item
  \textbf{Structured architecture}: Implement consistent hierarchical
  organization with explicit section boundaries. Each section should be
  self-contained for effective chunking.
\item
  \textbf{Speaker notes inclusion}: Embed speaker notes and
  supplementary explanatory content directly within the document to
  provide context that aids interpretation.
\end{enumerate}

Documents that follow the Web Content Accessibility Guidelines
(\citeproc{ref-wcag22}{World Wide Web Consortium 2024}) would be
well-suited not just for users with disabilities, but also for KMS.

\begin{figure}

\centering{

\includegraphics[width=1\linewidth,height=\textheight,keepaspectratio]{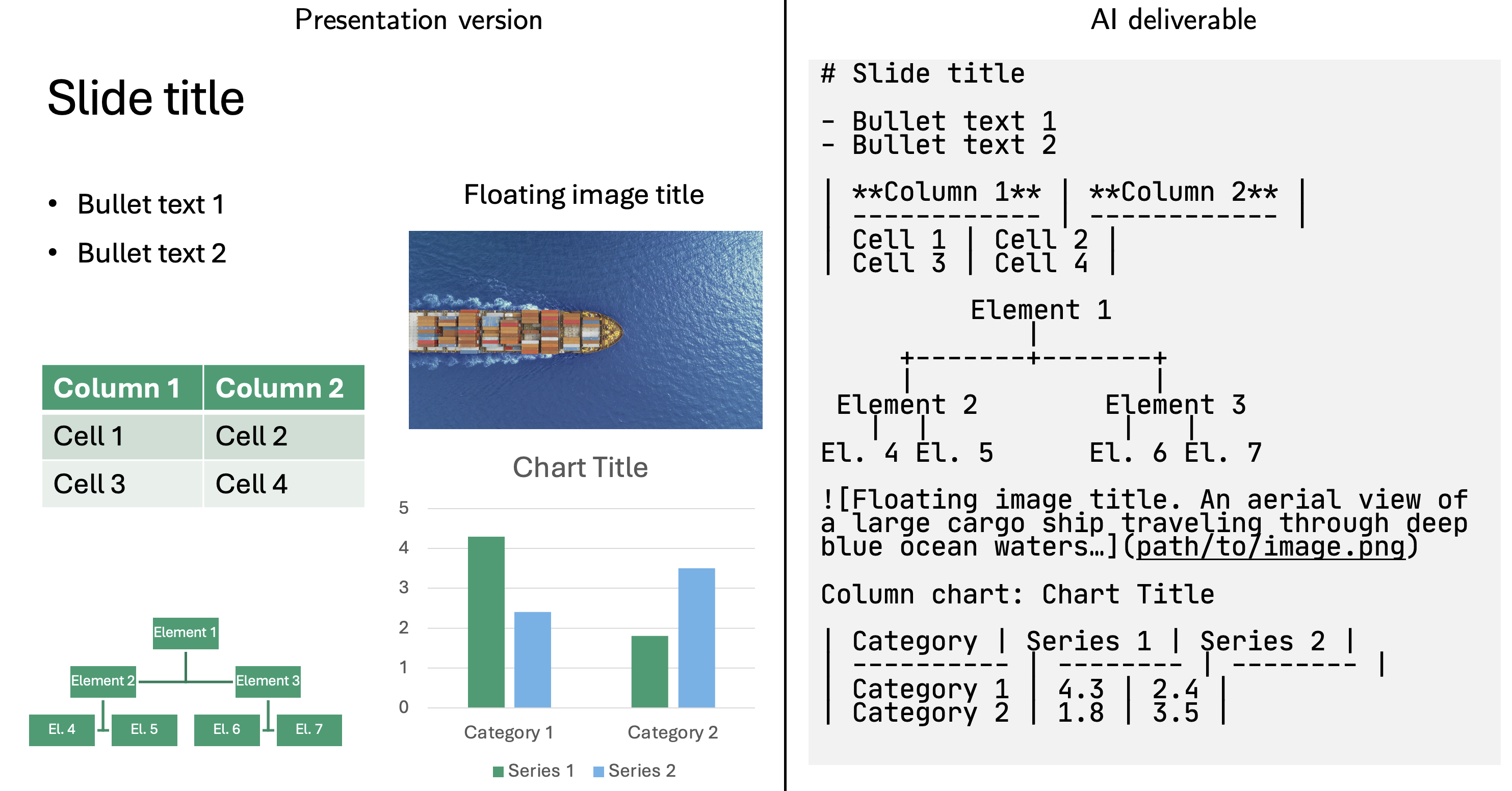}

}

\caption{\label{fig-special-purpose}An idealized Markdown representation
inspired by Zerox OCR. On the left is a PowerPoint slide, on the right
the Markdown representation as an AI-readable deliverable. The slide
features five layout elements: text, table, diagram, data chart, and
image. Each element is also represented in Markdown. The data chart is
converted to a table of numbers. The diagram is represented using ASCII
characters. The content of the image is described in a caption.}

\end{figure}%

A Markdown version of a document is a candidate for a special-purpose
deliverable, given the format's ubiquitous use with LLMs and support
from multiple conversion libraries. Markdown is a simple,
token-efficient plain-text format that supports formatting options like
image captions. The Markdown files produced by Zerox OCR push this
format's capabilities by transcribing diagrams into ASCII characters, an
approach that achieved the highest accuracy for diagram elements in our
benchmark. However, this method has its limitations: the transcription
is not standardized and may be inaccurate or infeasible for complex
diagrams. Markdown also generally lacks a structured way to represent
the spatial layout of elements, speaker notes, or color-coding from
tables and charts. Figure~\ref{fig-special-purpose} illustrates an
idealized output, showing how layout elements from a slide can be
represented in Markdown.

\subsection{Further research}\label{further-research}

Continued development of VLMs, layout models, and pipelines will advance
both the pipeline and pure vision approach. We expect that specific
feature gaps in pipeline libraries (like Docling's PPTX image
captioning) will be closed and bugs fixed.

Further end-to-end benchmarks and case studies with a focus on specific
layout elements and file formats are useful in this applied field. They
complement the existing QA and DLA benchmarks and provide a
comprehensive practitioner's view. Besides the technical side of
document understanding, there are organizational, cultural and
skill-related aspects to consider in the context of report preparation.

\section*{References}\label{references}
\addcontentsline{toc}{section}{References}

\phantomsection\label{refs}

\section*{Appendix}\label{appendix}
\addcontentsline{toc}{section}{Appendix}

\subsection*{Settings}\label{settings}
\addcontentsline{toc}{subsection}{Settings}

The LLM calls for conversion and answering use a temperature of 0.0. The
judge model does not support a temperature parameter. All other settings
were left at their default values.

\subsection*{Prompts}\label{prompts}
\addcontentsline{toc}{subsection}{Prompts}

\subsubsection*{Image captioning}\label{image-captioning}
\addcontentsline{toc}{subsubsection}{Image captioning}

\begin{verbatim}
Describe the image in detail. Extract numbers, text and everything else
required to answer questions about it. Do not use line breaks or
other formatting.
\end{verbatim}

This image captioning prompt is only used by Docling.

\subsubsection*{Question answering}\label{question-answering}
\addcontentsline{toc}{subsubsection}{Question answering}

\begin{verbatim}
You are a helpful assistant that answers questions based on provided context.

CONTEXT:
{report_content}

QUESTION:
{question}

Answer the question based solely on the provided context.
If the information isn't available in the context, say
"I don't have enough information to answer this question."
\end{verbatim}

\subsubsection*{Judging}\label{judging}
\addcontentsline{toc}{subsubsection}{Judging}

\begin{verbatim}
You are the judge for an AI system evaluation.

The AI was asked a question and provided an answer.
Your task is to check whether the answer matches the ground truth.

Correct answers:
- Contain all information from the ground truth
- May contain additional information, as long as it's not contradictory

Incorrect answers:
- Contradict the ground truth, even in parts
- Lack key information of the ground truth
- State that the question can't be answered

QUESTION:
{question}

GROUND TRUTH:
{ground_truth}

ANSWER:
{answer}
\end{verbatim}

\end{document}